%% file: acl_latex.tex
\pdfoutput=1

\documentclass[11pt]{article}

\usepackage[final]{acl}

\usepackage{times}
\usepackage{latexsym}

\usepackage[T1]{fontenc}

\usepackage[utf8]{inputenc}

\usepackage{microtype}

\usepackage{inconsolata}

\usepackage{enumerate}
\usepackage{amsthm}
\usepackage{graphicx}
\usepackage{epstopdf}
\usepackage{caption}
\usepackage{color}
\usepackage{enumitem}
\usepackage{array}
\usepackage{tabularx}
\usepackage{multirow}
\usepackage{bm}
\usepackage{booktabs}

\usepackage{tcolorbox}
\usepackage{caption}
\usepackage{subcaption}
\usepackage{listings}
\usepackage{wrapfig}
\usepackage{bbding}
\tcbuselibrary{skins}
\usepackage{colortbl}
\usepackage{fontawesome}

\usepackage{authblk}
\usepackage{amsmath}

\usepackage{etoolbox}
\preto{\abstractkeywords}{\nolinenumbers} 

\title{Do LLMs Really Think Step-by-step In Implicit Reasoning?}

\author[a]{Yijiong Yu}
\affil[a]{Tsinghua University}

\begin{document}
\maketitle
\begin{abstract}

It has been well-known that Chain-of-Thought can remarkably enhance LLMs' performance on complex tasks. However, because it also introduces slower inference speeds and higher computational costs, many researches have attempted to use implicit CoT, which does not need LLMs to explicitly generate the intermediate steps. However, the invisible reasoning process leaves us a doubt that, can implicit CoT really be equal to explicit CoT? Therefore, in this study, we address this question through experiments. We probe the information of intermediate steps from the model's hidden states when it is either trained or prompted to perform implicit CoT. The results surprisingly indicate that when prompted, LLMs hardly think about intermediate steps, suggesting they may just rely on experience rather than strict step-by-step reasoning. But when trained, they indeed calculate intermediate steps. Moreover, in both situations, we find the effect of using implicit CoT is susceptible to the format of the problem, reaffirming the current deficiency of implicit CoT.

\end{abstract}

\section{Introduction}

Advancements in Large Language Models (LLMs) have unveiled unprecedented capabilities in handling complex reasoning tasks. Chain-of-Thought (CoT) prompting \cite{wei_chain--thought_2023,yu_towards_2023}, in particular, has demonstrated substantial improvements in the reasoning abilities of LLMs by explicitly mapping out intermediate reasoning steps. Recent products like OpenAI o1 \cite{qin_o1_2024} further demonstrate the power of CoT. 

However, the CoT approach, despite its efficacy, it notably incurs slower inference speeds and higher computational costs. These drawbacks have spurred some researches on alternative reasoning methodologies that bypass the explicit generation of intermediate tokens, leveraging the model's inherent ``vertical'' reasoning capabilities through its internal processing layers. For example, \cite{deng_explicit_2024} gradually remove the intermediate steps and fine-tune the model to let it learn implicit CoT, and \cite{deng_implicit_2023} train a emulator which emulate the intermediate states in CoT reasoning and train a student model to generate answers from these implicit states. This form of reasoning does not need to output intermediate results as tokens, called implicit reasoning or vertical reasoning, which contrasts with the ``horizontal'' reasoning, i.e. typical CoT. We show the difference between explicit CoT and implicit CoT with figures in Appendix \ref{app:dif}

Although the concept of ``implicit CoT (reasoning)'' is rarely directly mentioned, in many scenarios that require low latency, users usually ask LLMs to output the final answer directly, which actually has forced LLMs to adopt the same answering way as implicit CoT.

Despite the theoretical appeal of implicit reasoning as a more efficient alternative to traditional CoT methods, our empirical evidence suggests the performance of implicit CoT still lag behind explicit CoT. Moreover, though some previous researches have confirmed the concept of implicit reasoning and attempted to analyze its process and efficacy \cite{yang_multihop_2024,wang_grokked_2024,allen-zhu_physics_2024}, they usually more focus on using knowledge-based problems to examine whether LLMs can recall their parametric knowledge during implicit reasoning, instead of investigating problems with more steps such as arithmetic. So far, there is still no clear and widely accepted conclusion on the rationale of implicit reasoning.

This situation makes us wonder fundamental questions about the nature of the implicit reasoning, such as ``Are LLMs doing the same thing in the processes of implicit and explicit CoT?'' and ``Can the hidden, internal and layer-by-layer processing truly serve as an equivalent to explicit CoT reasoning?'' To answer these, our study designs experiments aimed at uncovering the implicit reasoning processes, specifically targeting the process of handling multi-step arithmetic problems without resorting to outputting explicit intermediate steps.

In our experiment, we leverage 2 models which may have the ability of implicit CoT, a powerful generic model, Qwen2.5-72B-Instruct \cite{team_qwen25_2024}, and a special Mistral \cite{jiang_mistral_2023} model trained by Internalizing CoT step by step \cite{deng_explicit_2024}. We let them tackle simple arithmetic problems that are easily solvable via typical CoT reasoning \cite{ye_physics_2024}. However, we force the model to direct give the answer without outputting steps, so that we can examine whether these tasks can be addressed through implicit reasoning and how implicit reasoning happens. 

The arithmetic problems has controllable number of reasoning steps, with each intermediate result being known. By investigating the hidden states associated with the final token of the given problem statement across layers and employing a simple linear classifier to probe those intermediate results, we aim to find out if the model really calculates the intermediate results in its implicit thinking process.

The experiment results are surprising: if being prompted to perform implicit CoT, the model hardly calculates the intermediate results when there are more than 1 intermediate step, despite it can often give the correct answer of the multi-step problem. However, if being trained to, the situation is exactly the opposite: it indeed calculates step by step.

Moreover, we conduct another experiment to test if the latter situation (trained) of implicit CoT is more stable and robust than the former (prompted), because it truly thinks step by step. Surprisingly, after slightly modifying the problem format without even increase its difficulty, the efficacy of implicit CoT degrades severely in both situations, which suggests implicit CoT is faster but less reliable, whether the model is prompted or trained. 

In conclusion, we think implicit CoT can indeed simulate the behaviors of explicit CoT to some extent, but it cannot substitute explicit CoT, because its rationale is quite different and its efficacy in solving practical problems is still far from satisfactory. Our study provides critical insights into the mechanics of implicit reasoning and emphasizes the ongoing necessity for explicit CoT methodologies in enhancing LLMs ability on complex tasks.


\section{Approach}
\subsection{Expriment Design}
\label{setting}
To present the reasoning steps clearly, we adopt simple multi-step arithmetic problems with only addition and subtraction. Usually, when given such problems, modern LLMs will automatically use a CoT manner to address them. To investigate the process of implicit reasoning, we use prompt to force the model to give the answer without using CoT. Therefore, an example of our prompt, which is a 5-step problem, is as follows:

\begin{tcolorbox}[colback=white]
\small{{\sffamily
$E = 8 ;$\\
$D = E - 5 ;$\\
$C = D + 2 ;$\\
$B = C + 5 ;$\\
$A = B - 1 ;$\\

Question: What is the value of A? You must answer directly with A=xxx. \\

Answer: A=
}}
\end{tcolorbox}

If the model is already trained to internalize CoT, we do not need to add additional instructions. So the prompt (following the format of \cite{deng_explicit_2024}) is:
\begin{tcolorbox}[colback=white]
\small{{\sffamily
$E = 8 ;$\\
$D = E - 5 ;$\\
$C = D + 2 ;$\\
$B = C + 5 ;$\\
$A = B - 1 ;$\\
$A=?$\\
</s></s>\#\#\#\#
}}
\end{tcolorbox}

We randomly change the value in the problem to generate 2000 different samples, and each intermediate result is recorded. For example, the intermediate results of the above example should be $[8, 3, 5, 10, 9]$, i.e. the corresponding value of E, D, C, B and A. The result of the last step is the final answer. 

\begin{figure*}[htb]
    \centering
    \subfloat[Qwen 3-step]{\includegraphics[width=0.33\linewidth]{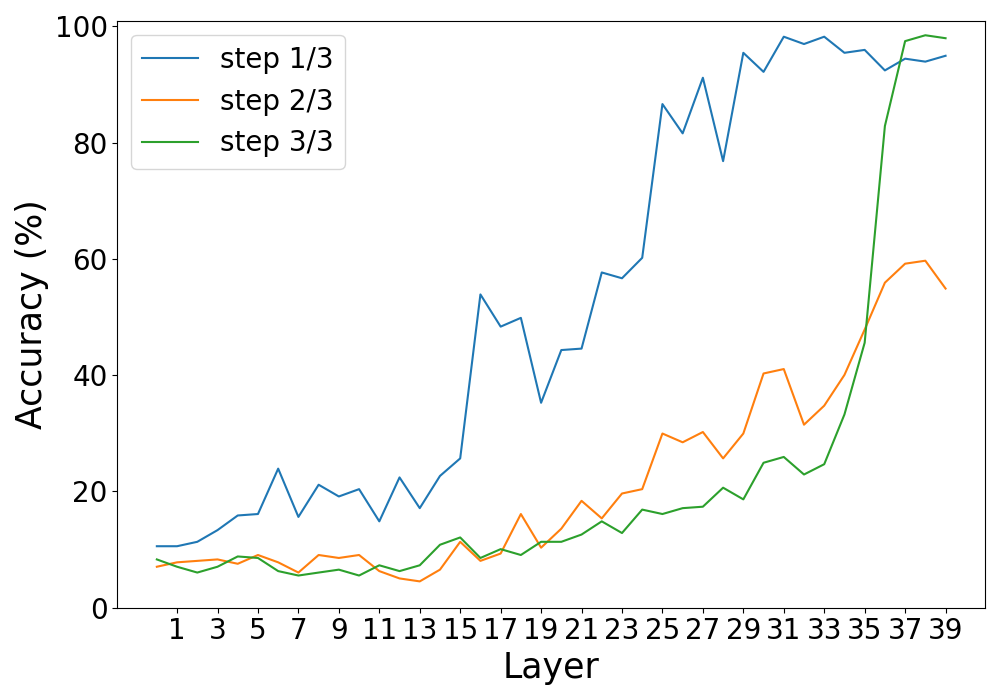}}
    \subfloat[Qwen 4-step]{\includegraphics[width=0.33\linewidth]{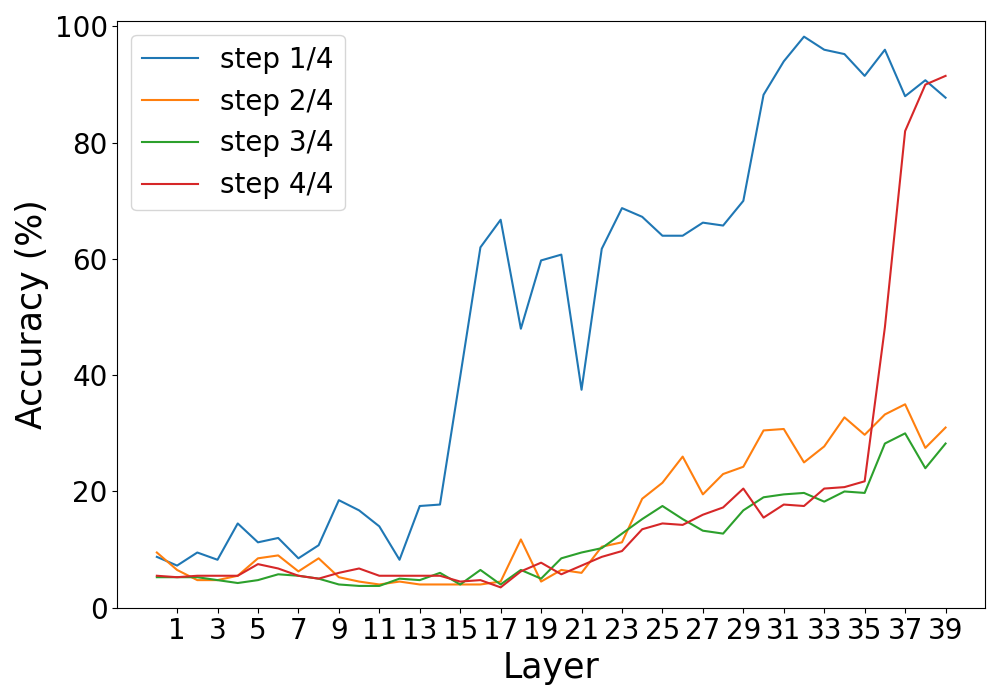}}
    \subfloat[Qwen 5-step]{\includegraphics[width=0.33\linewidth]{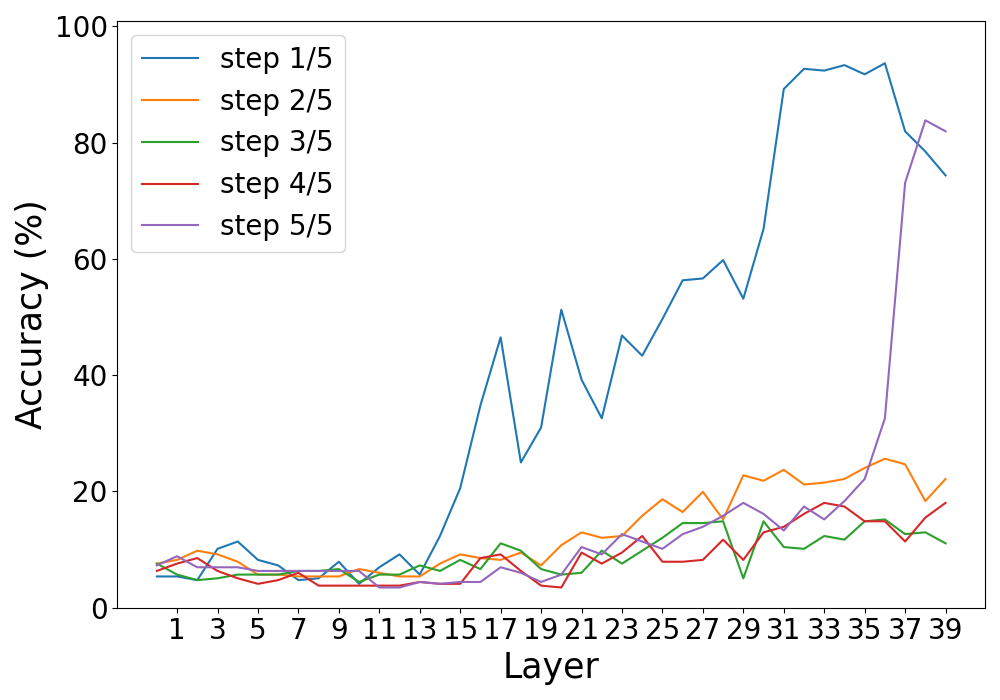}}
    
    \subfloat[Mistral 3-step]{\includegraphics[width=0.33\linewidth]{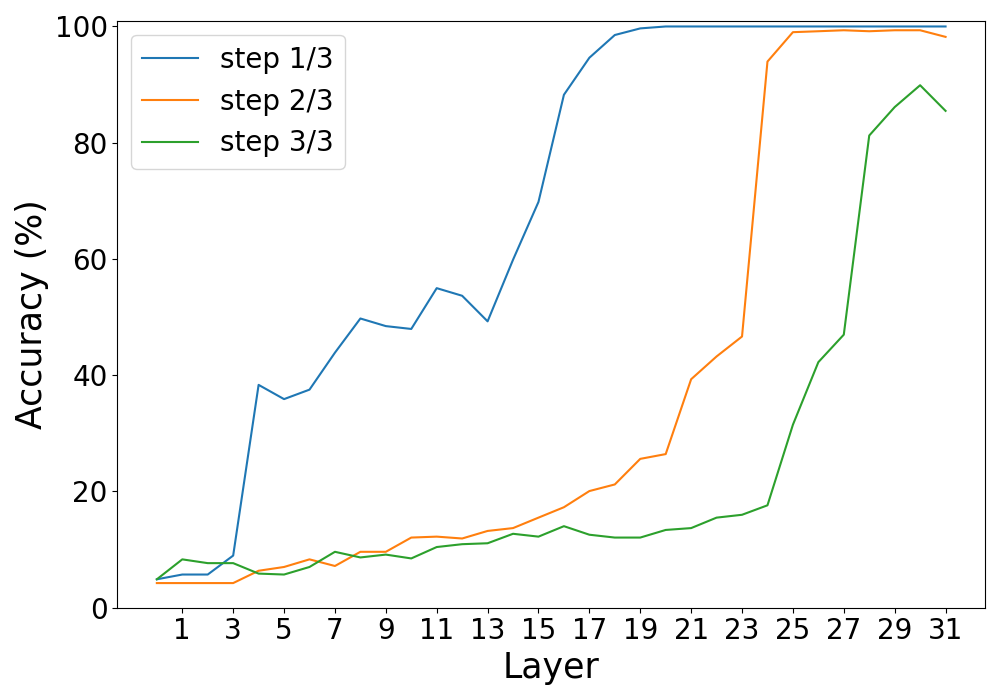}}
    \subfloat[Mistral 4-step]{\includegraphics[width=0.33\linewidth]{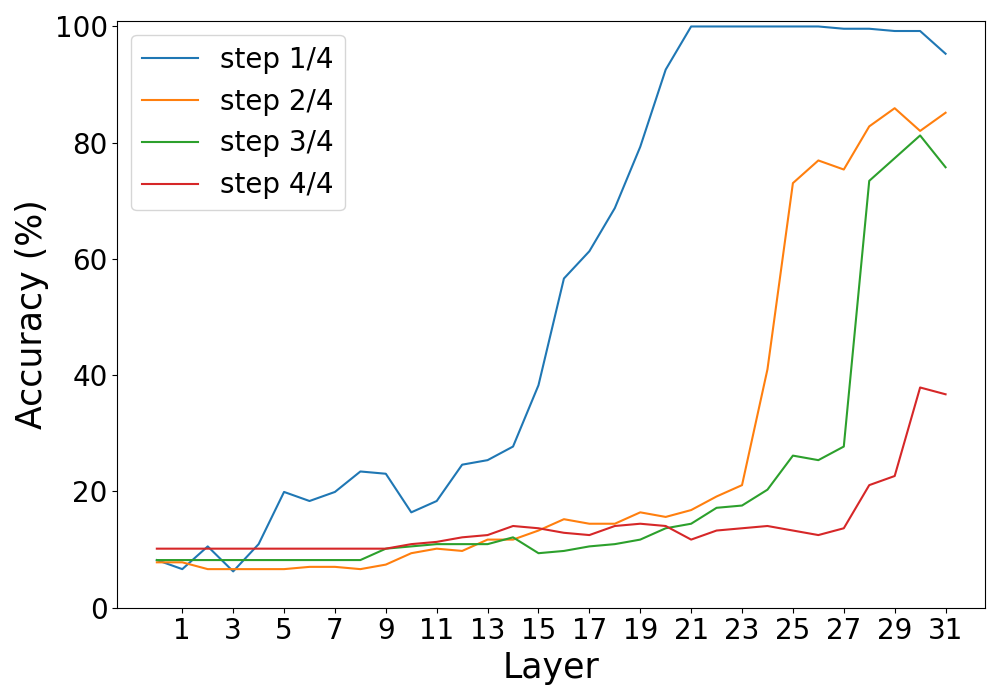}}
    \subfloat[Mistral 5-step]{\includegraphics[width=0.33\linewidth]{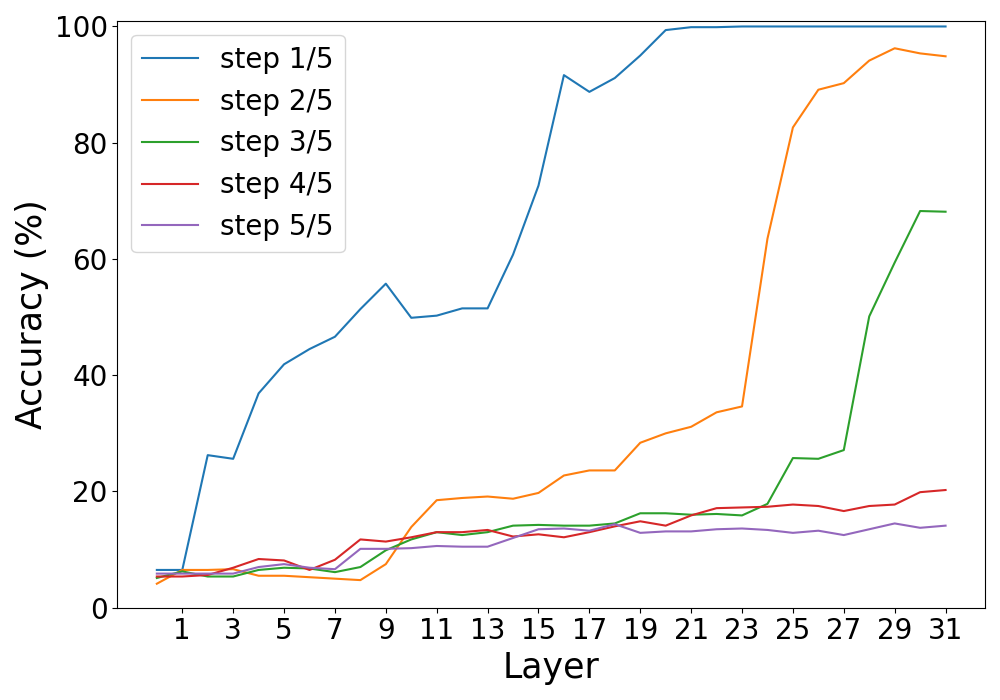}}

    \caption{The accuracy of probing the results of each step in multi-step arithmetic problems. Qwen2.5 represents we prompt Qwen2.5-72b-instruct to perform implicit CoT. Mistral represents the mistral-7b-v0.1 model which has been trained to internalize CoT.}
    \label{fig:probe}
\end{figure*}

The model will direct output the answer after our prompt, thus we take the last token of the prompt as our main research object and record its hidden states of each layers. Then, we adopt a typical linear probing method using an 1-layer MLP, to predict each of the intermediate results from the hidden states of each layer. We control the values of all intermediate results and the final answer within the range of 0-19 (inlucding 0 and 19), so that the probe is a 20-class classifier (each value corresponds to one class). More details about why we choose this range and how we generate samples are in Appendix \ref{app:data}.

We use 1600 samples to train the classifier for 8 epochs and 400 samples for testing its accuracy. We train an individual classifier for each layer, using the results of each step of the problem as the labels. If the classifier corresponding to the $k$-th layer and the $n$-th step shows high accuracy in the test set, it represents the model has calculated the result of the $n$-th step in the $k$-th layer's hidden states.

In the first situation (prompted), we choose a large model, Qwen2.5-72B-Instruct \cite{team_qwen25_2024}, to perform implicit reasoning, because we find small 7B-level models without specifically trained for implicit CoT can hardly answer a multi-step problem correctly without explicit CoT, while a 70B level model can achieve an accuracy of over 50\% even when 5-step, demonstrating its ability to use implicit CoT to solve problems. Because the 72B model has 80 layers, to reduce the computing cost, we average the hidden states across every 2 consecutive layers (so it is equivalent to a 40-layer model), and the model is quantized to 4-bit. 


In the second situation (trained), we choose mistral-7b-internal-CoT \cite{deng_explicit_2024}, which has been trained by a multi-stage method to internalize the step-by-step CoT process on GSM8k \cite{cobbe_gsm8k_2021}. It has the ability to directly output the answer after a relatively simple math problem. It has 32 layers and we record its original hidden states of each layer without compression.

By default, in all generation experiments, the temperature is set to 0.

\subsection{Results of Probing Intermediate Steps}

From the results of the probing experiments in Figure \ref{fig:probe}, we can draw the following conclusions:

(1) When prompted to perform implicit CoT, the results of the first step and the last step can always be probed successfully in the back layers, indicating the model does memorize the input value (i.e. the result of the first step) and does conceive the final answer (i.e. the result of the last step). However, the results of intermediate steps can hardly be detected, especially in the problem with more steps. It looks that the curve of the last step just surges in the last layers, even without waiting for the processing of the previous step. Therefore, we speculate that it may just rely on rich experience to guess the answer correctly.

(2) For the model which has been trained to perform implicit CoT, we can indeed observe a process of calculating intermediate results step by step. The curve of each step starts to rise right after the curve of the previous step reaches a level, which is consistent with a natural CoT process. 

(3) However, for the trained model, when the problem's steps are getting more, the process of calculating each step is postponed to later layers. For example, in the 3-step problem, the curve of the 2rd step starts to rise rapidly at layer 20, while in the 4-step problem, it is at layer 23. As a result, when facing the 5-step problem, although the model has attempted to calculate the beginning steps, there are no layer left to process the remaining steps (the 4th and 5th). This indicates the capacity of implicit CoT is very limited by the number of layers.

This contrast indicates that, although their external appearances seem the same, the internal processes of implicit CoT are completely different when prompted or trained. Internalizing CoT step by step can indeed make model perform ``vertical'' reasoning, but conventional models seem unable to truly think step by step, even if answering correctly without outputting intermediate steps.

\subsection{Result of Modifying the Problem Presentation}
We have found the internal process of trained or prompted implicit CoT is different. Therefore, we want to find out whether they also have different external performance. In our initial speculation, the trained way of implicit CoT should be more robust facing various problems. In this section, we test the accuracy of solving arithmetic problems by implicit CoT in both situations, when the problems require more steps, and the problem format is changed with the actual difficulty remaining unchanged.

To change the problem form, we modify the problem by 2 ways: 1. reversing the order of the equations; 2. dividing all values by 10. Thus we obtain 2 additional types of problem presentations. For humans or LLMs performing conventional CoT, such modifications hardly increase any difficulty of the problem, because the reasoning steps are almost not changed at all. The examples of the modified problem are as follows:

\begin{tcolorbox}[colback=white, title=Reverse]
\small{{\sffamily
$A = B - 1 ;$\\
$B = C + 5 ;$\\
$C = D + 2 ;$\\
$D = E - 5 ;$\\
$E = 8 ;$
}}
\end{tcolorbox}

\begin{tcolorbox}[colback=white, title=Scale]
\small{{\sffamily
$E = 0.8 ;$\\
$D = E - 0.5 ;$\\
$C = D + 0.2 ;$\\
$B = C + 0.5 ;$\\
$A = B - 0.1 ;$
}}
\end{tcolorbox}


Besides testing implicit CoT (prompted or trained), we also test other instruction-tuned models (we tested the 1.5b, 7b, 72b model of Qwen2.5 series) with explicit CoT (adding ``let's think step by step'' to let it output reasoning steps). The results show that they always achieve nearly 100\% accuracy with explicit CoT, regardless of the steps or format of the problem. However, if prompted not to output steps, only the 72b model can reach an accuracy over 50\% in a 5-step problem.

By contrast, from the results of implicit CoT in Table \ref{tab:accuracy}, we can clearly see that, compare to the original problems, the modified problems significantly degrade the performance for both situations (prompted or trained), especially when the order of equations is reversed. For the trained way, despite it truly thinks step by step, the accuracy even drops more significantly. The probing results in Appendix \ref{app:rev} show that reversing makes mistral-internal-CoT unable to calculate the intermediate results at all. This finding indicates implicit CoT is usually much less reliable and stable. 


    


\begin{table}[htbp]
\centering

\begin{tabular}{c|c|ccc}
\toprule
Model & Step & Original & Reversed & Scaled \\
\midrule
\multirow{3}{*}{Mistral} & 3 & 78.2 & 13.7 & 10.0 \\
& 4 & 31.9 & 4.7 & 15.2 \\
& 5 & 4.8 & 1.8 & 5.4 \\
\midrule
\multirow{3}{*}{Qwen} & 3 & 99.8 & 89.0 & 90.1 \\
& 4 & 83.2 & 41.9 & 80.5 \\
& 5 & 65.2 & 14.9 & 59.1 \\
\bottomrule
\end{tabular}
\caption{The accuracy (\%) of Qwen (prompted) and Mistral (trained) under different problem formats and steps using implicit CoT.}

\label{tab:accuracy}
\end{table}

\section{Conclusion}

In this study, we investigate the mechanism of implicit CoT, and get a non-trivial finding that, unlike some previous studies which envisioned implicit reasoning as a substitute for explicit reasoning, implicit reasoning cannot be on par with explicit reasoning methods because it may not follow a step-by-step process in all problems. Moreover, it is susceptible to the input format, which makes it less reliable. This finding remind us that implicit CoT still needs to be improved to ensure accuracy and robustness. When you ask LLMs to give the answer directly, you should know that it has not actually undergone a real reasoning process. Using explicit CoT may still be the most feasible method to further propel the capabilities of LLMs at present.

\section{Limitations}
We only test one type of reasoning problems, the arithmetic problem. That is because for other types of problems, the values of the intermediate steps and the final answer are hard to control strictly.

Some newly proposed methods \cite{hao_latent_2024} may improve the effect of implicit CoT, but we have not yet experiment with them.


\bibliography{custom2}
\appendix

\input{appendix}

\end{document}

%% file: appendix.tex
\begin{figure*}[htb]
    \centering
    \includegraphics[width=0.85\linewidth]{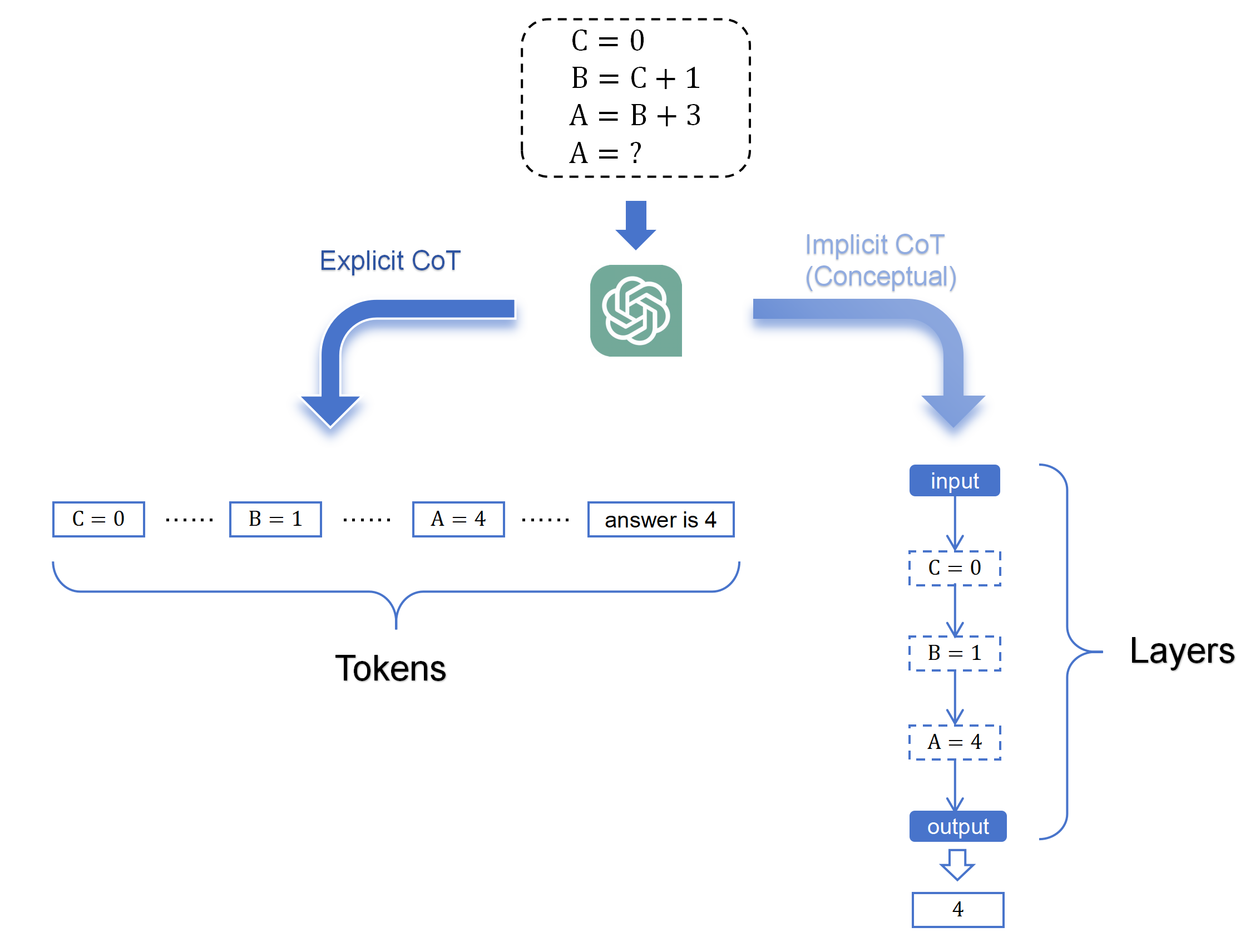}
    \caption{The examples of explicit CoT and implicit CoT. Explicit CoT is commonly used, which is completed by step-by-step output tokens. The process of implicit CoT is just a hypothetical or conceptual situation, which could be a layer-by-layer way.}
    \label{fig:implicit cot}
\end{figure*}

\section{Difference between implicit and explicit CoT}
\label{app:dif}

Figure \ref{fig:implicit cot} show the difference between implicit and explicit CoT with an example. In explicit CoT, LLM output tokens to represent the reasoning process, while in implicit CoT, LLM directly output the answer with the reasoning process is done in its layer-by-layer forward process.

\section{Data Generation}
\label{app:data}
In the probing experiment, to train a fair probe model, we need to keep the number of each label is nearly the same, i.e., they obey discrete uniform distribution. 

In some our early probing experiments, we found LLMs are worse at handling negative numbers than positive numbers: the probing accuracy of the intermediate result of a negative number is generally much lower than that of a positive number. Therefore, for fairness, we make all the intermediate results $R$ random positive numbers or 0. Specifically, we make it obey this distribution:

$$P(R_i=x)=1/(M+1), x\in [0,M], M\in N_+$$

Here $R_i$ represents the result of the $i$-th step. $R_0$ represents the initial value. And we let $A_i$ represent the addend in the $i$-th step.

To generate uniform distributed $R_i$, in practice, we actually first generate the intermediate results using \textit{random.randint()}, and then calculate $A_i$ by $A_i=R_i-R_{i-1}$.

We set $M$ to 19 in our experiment, because we find setting $M$ to 10 will make the data lack diversity, while a too large $M$ will slow the training of the probe.


\begin{figure}[htb]
    \centering
    \includegraphics[width=0.8\linewidth]{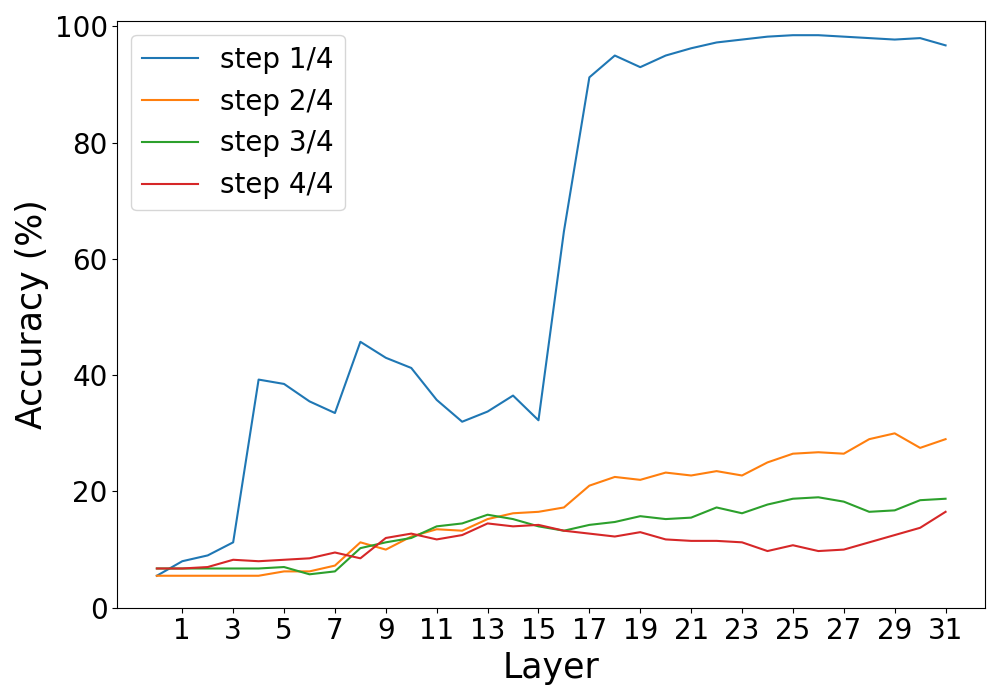}
    \caption{The accuracy of probing the results of each step in multi-step arithmetic problems from the hidden states of mistral-internal-CoT, when the order of the equations is reversed.}
    \label{fig:rev}
\end{figure}

\section{Additional Results}
\label{app:rev}
When the order of the equations in the problem is reversed, the probing results is shown in Figure \ref{fig:rev}. We can see that the intermediate results as well as the final result cannot be calculated, while only the initial value can be probed. This results demonstrate current implicit-CoT methods are susceptible to the format of the problem.